\documentclass[letterpaper, 10 pt, conference]{ieeeconf}  

\IEEEoverridecommandlockouts                              

\usepackage{times}

\usepackage{cite}

\usepackage{multicol}

\usepackage{xcolor} 
\usepackage{graphicx}

\usepackage{booktabs}
\usepackage{multirow}
\usepackage{algorithmic}
\usepackage{algorithm}
\usepackage{amsmath}
\usepackage{amssymb}
\usepackage{etoolbox}

\usepackage[acronym]{glossaries}
\glsdisablehyper   

\usepackage[colorlinks=true, citecolor=blue]{hyperref}

\usepackage{siunitx}

\usepackage[font=small, labelfont=bf, tableposition=top]{caption}
\usepackage{subcaption}

\usepackage{array}

\usepackage{float}  
\usepackage{stfloats}

\newcolumntype{P}[1]{>{\raggedright\arraybackslash}p{#1}}

\makeatletter
\newcommand\fs@betterruled{%
  \def\@fs@cfont{\bfseries}\let\@fs@capt\floatc@ruled
  \def\@fs@pre{\vspace*{5pt}\hrule height.8pt depth0pt \kern2pt}%
  \def\@fs@post{\kern2pt\hrule\relax}%
  \def\@fs@mid{\kern2pt\hrule\kern2pt}%
  \let\@fs@iftopcapt\iftrue}
\floatstyle{betterruled}
\restylefloat{algorithm}
\makeatother

\usepackage{etoolbox}
\newcommand{\insertfig}{\includegraphics[width=0.98\linewidth]
{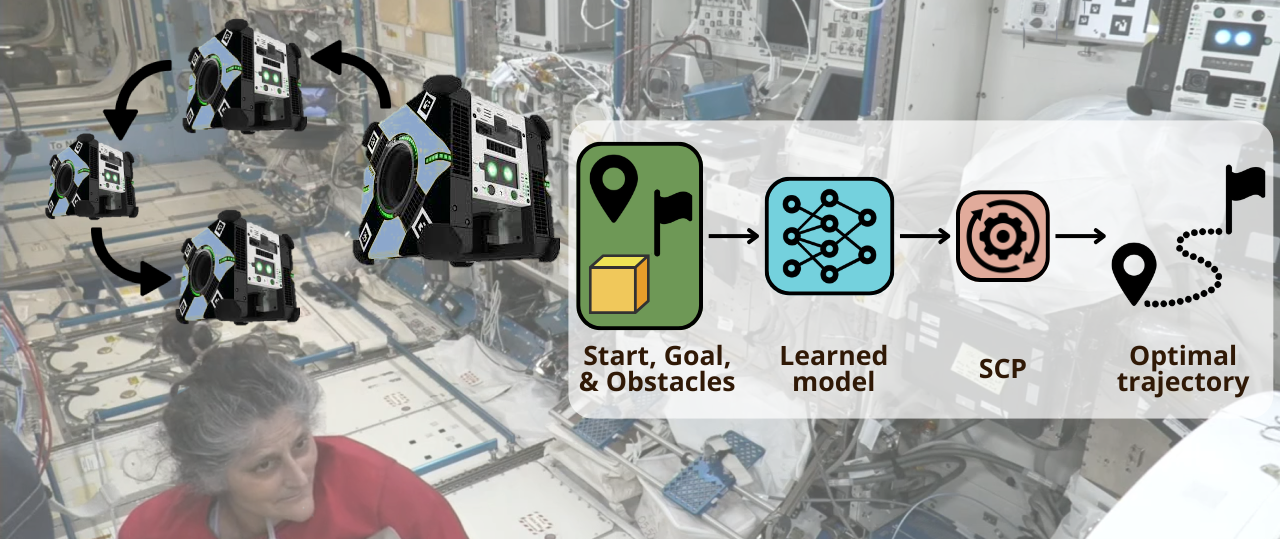}
\captionof{figure}
{In this work, we present results from the first in-space demonstration of machine learning-based warm starts for accelerating trajectory optimization during experiments conducted onboard the International Space Station with the Astrobee free-flying robot.}
\label{fig:title_figure}
}

\makeatletter
\apptocmd{\@maketitle}{\centering\insertfig\vspace{-0.5em}}{}{}
\makeatother

\makeatletter
\def\@citex[#1]#2{%
  \let\@citea\@empty
  \@cite{\@for\@citeb:=#2\do
    {\@citea\def\@citea{,\penalty\@m\ }%
     \edef\@citeb{\expandafter\@firstofone\@citeb}%
     \if@filesw\immediate\write\@auxout{\string\citation{\@citeb}}\fi
     \@ifundefined{b@\@citeb}{\hbox{\reset@font\bfseries ?}%
       \G@refundefinedtrue
       \@latex@warning
         {Citation `\@citeb' on page \thepage \space undefined}}%
       {\hyper@natlinkstart{\@citeb}\@cite@ofmt{%
         \csname b@\@citeb\endcsname}\hyper@natlinkend}}}{#1}}

\AtBeginDocument{%
  \let\old@bibitem\@bibitem
  \renewcommand{\@bibitem}[1]{%
    \old@bibitem{#1}%
    \Hy@raisedlink{\hyper@anchorstart{cite.#1}\hyper@anchorend}%
  }%
}
\makeatother

\newacronym{mpc}{MPC}{model predictive control}
\newacronym{rl}{RL}{reinforcement learning}
\newacronym{ai}{AI}{artificial intelligence}
\newacronym{ml}{ML}{machine learning}

\newacronym{iss}{ISS}{International Space Station}
\newacronym{gusto}{GuSTO}{Guaranteed Sequential Trajectory Optimization}

\newacronym[longplural={guidance, navigation, \& controls},shortplural={GNC}]{gnc}{GNC}{guidance, navigation, \& control}

\newacronym{scp}{SCP}{sequential convex programming}
\newacronym[longplural={quadratic programs},shortplural={QPs}]{qp}{QP}{quadratic program}

\renewcommand{\baselinestretch}{0.89275}

\usepackage[font=small, labelfont=bf, tableposition=top]{caption}

\usepackage{subcaption}

\graphicspath{{./fig/}}

\newcommand{\eg}{{e.g.}}
\newcommand{\ie}{{i.e.}}

\begin{document}

\title{\LARGE \bf
Deep Learning Warm Starts for Trajectory Optimization \\on the International Space Station
}


\author{Somrita Banerjee$^{1}$, Abhishek Cauligi$^{2}$, and Marco Pavone$^{3}$
\thanks{$^{1}$Somrita Banerjee is a Machine Learning Researcher at Apple. Work completed during PhD at Stanford University,
        Stanford, CA 94305, USA.
        {\tt\small somritabanerjee@gmail.com}}%
\thanks{$^{2}$Abhishek Cauligi is an Assistant Professor in the Department of Mechanical Engineering, Johns Hopkins University,
        Baltimore, MD 21218, USA. {\tt\small cauligi@jhu.edu}}%
\thanks{$^{3}$Marco Pavone is an Associate Professor in the Department of Aeronautics and Astronautics, Stanford University,
        Stanford, CA 94305, USA. {\tt\small pavone@stanford.edu}}%
}



%

\typeout{>>> Figure counter before hero figure: \thefigure}

\maketitle
\setcounter{figure}{1}

\typeout{>>> Figure counter after hero figure: \thefigure}

\begin{abstract}
Trajectory optimization is a cornerstone of modern robot autonomy, enabling systems to compute trajectories and controls in real-time while respecting safety and physical constraints.
However, it has seen limited usage in spaceflight applications due to its heavy computational demands that exceed the capability of most flight computers.
In this work, we provide results on the first in-space demonstration of using machine learning-based warm starts for accelerating trajectory optimization for the Astrobee free-flying robot onboard the International Space Station (ISS).
We formulate a data-driven optimal control approach that trains a neural network to learn the structure of the trajectory generation problem being solved using sequential convex programming (SCP).
Onboard, this trained neural network predicts solutions for the trajectory generation problem and relies on using the SCP solver to enforce safety constraints for the system.
Our trained network reduces the number of solver iterations required for convergence in cases including rotational dynamics by 60\% and in cases with obstacles drawn from the training distribution of the warm start model by 50\%.
This work represents a significant milestone in the use of learning-based control for spaceflight applications and a stepping stone for future advances in the use of machine learning for autonomous guidance, navigation, \& control.
\end{abstract}

\IEEEpeerreviewmaketitle

\section{Introduction}
Trajectory generation is a hallmark problem of aerospace~\gls{gnc} and involves the computation of a state and control trajectory that satisfies system dynamics and mission operational constraints.
Traditionally, trajectory generation for space systems is carried out with significant ground-in-the-loop involvement, where mission designers construct the trajectory using large-scale nonlinear programming~\cite{Betts1998,Betts2010} or primer-vector techniques~\cite{HandelsmanLion1967,Russell2007}.
However, with the advent of a new era of spaceflight involving an increasing number of missions in cislunar space~\cite{HolzingerChowEtAl2021} and a burgeoning interest for in-space servicing capabilities, the current state-of-practice in trajectory generation for space systems falls short in fulfilling the needs of upcoming missions.
In particular, there is a pressing need to be able to compute trajectories autonomously onboard spacecraft and allow for scaling to an increasing number of missions without incurring significant operational costs.

A key challenge in enabling onboard trajectory generation is that the nonlinear optimization algorithms used to formulate and solve these problems remain too computationally expensive for resource-constrained flight computers.
Indeed, despite recent advances in embedded convex optimization solvers~\cite{ErenPrachEtAl2017,LiuLuEtAl2017,MalyutaYuEtAl2021,MalyutaEtAl2022}, most practical problems of interest in spacecraft applications fall under the class of non-convex optimization problems for which finding solutions remains even more computationally expensive.
In this work, we draw inspiration from data-driven optimal control, a nascent area of research that has emerged using machine learning to quickly synthesize solutions for the trajectory generation problem, to bridge this computational gap while providing the necessary safety guarantees for the system.

We present experimental results on the use of data-driven optimal control for the Astrobee free-flying robot.
The Astrobee robot is a guest science platform operating onboard the International Space Station~\cite{BualatSmithEtAl2018} and serves as a proving ground for new technologies developed for spacecraft robotic applications.
The trajectory generation problem for Astrobee entails solving a highly non-convex optimization problem and for which a new class of trajectory optimization solvers using~\gls{scp} have emerged to successfully tackle.
We showcase how data-driven optimal control techniques allow for the use of such~\gls{scp} algorithms for real-time, onboard for generating trajectories that satisfy system constraints while safely avoiding obstacles.

{\em Statement of Contributions: }
Our work presents experimental results from the first flight demonstration of data-driven optimal control used to generate trajectories for the Astrobee robot.
We use the~\gls{gusto} algorithm~\cite{BonalliCauligiEtAl2019,BonalliBylardEtAl2019} to solve the free-flyer trajectory optimization problem.
We then train a neural network to learn the mapping between problem inputs and~\gls{gusto}-generated solutions.
To create training data, we run~\gls{gusto} on randomized motion plans using a simulator of the~\gls{iss} and Astrobee dynamics.
The trained neural network is integrated into the Astrobee flight software~\cite{Astrobee} and was tested during an experimental session on board the~\gls{iss} on February 13, 2025.

The primary contributions of this paper include:
\begin{enumerate}
    \item We provide experimental results with the Astrobee free-flying robot conducted on the~\gls{iss} in February 2025, which represents the first use of machine learning-based control in the~\gls{iss} microgravity environment and, to the best of the authors' knowledge, for a free-flying robot.
    \item Our learned warm starts significantly reduce the number of solver iterations required for convergence in complex, non-convex planning tasks, including a 60\% reduction in iterations required for convergence in scenarios involving rotational dynamics, and a 50\% reduction in cases with obstacle configurations sampled from the warm start model's training distribution.
    \item As a result of these tests, we advance the Technology Readiness Level (TRL) of deep learning-based trajectory optimization from TRL-3 (experiments from ground based benchmarks) to TRL-5 (testing in a relevant environment).
\end{enumerate}

\begin{figure}[t]
  \centering
  \includegraphics[width=0.49\textwidth, trim=0 0 0 0, clip]{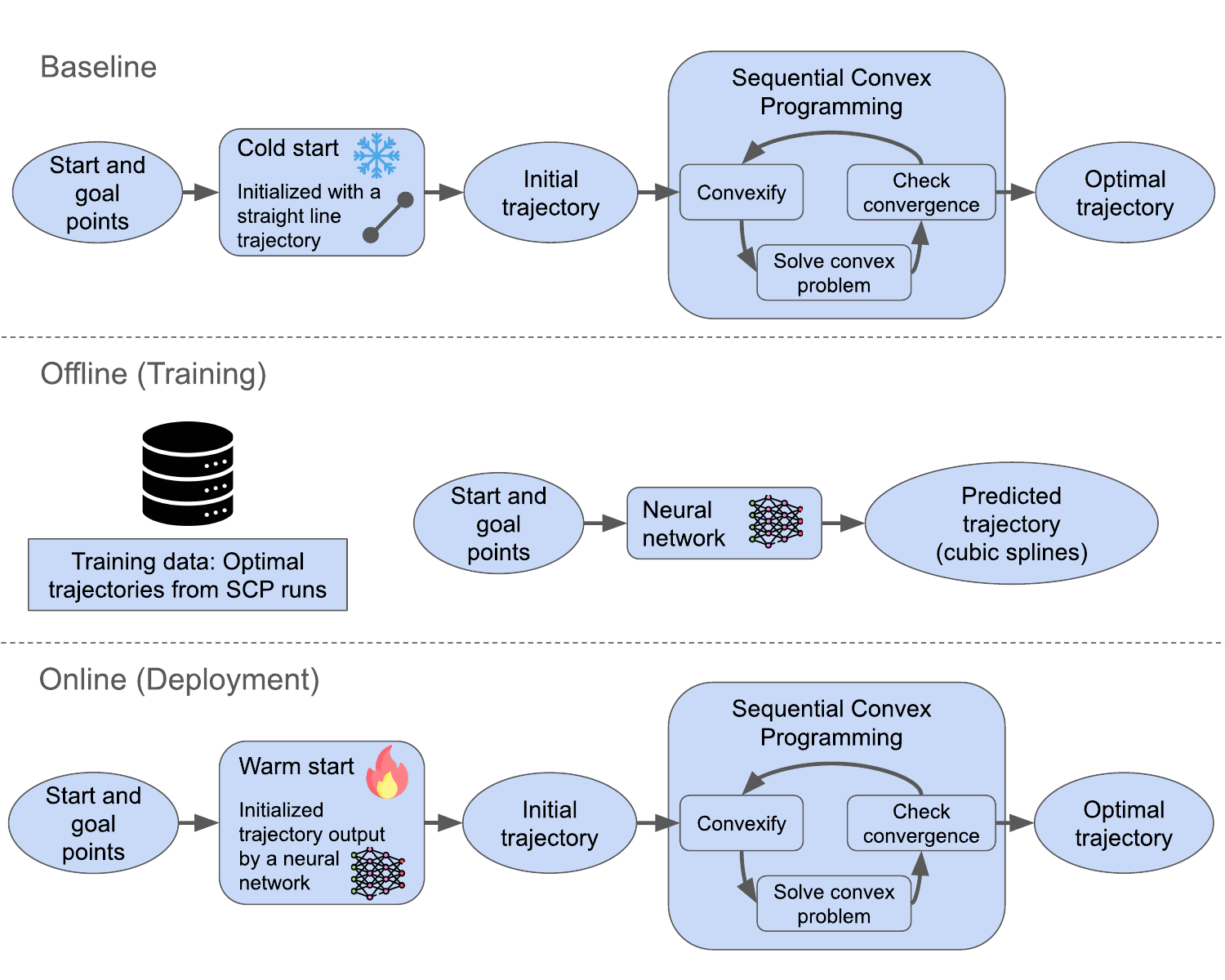}
  \caption{\textbf{Overview of our learning-based warm-starting approach for accelerating trajectory optimization.} The baseline method uses Sequential Convex Programming (SCP) initialized with a cold start (\ie{}, a straight-line trajectory). In the offline phase, a neural network is trained on solved planning problems to predict effective warm starts. At deployment (online), the learned warm start is used to initialize SCP, resulting in faster convergence to an optimal trajectory.}
  \label{fig:flowchart}
\end{figure}

\section{Related Work}~\label{sec:related_work}
The application of machine learning to solve planning and control problems for aerospace systems has been an extensive area of study~\cite{IzzoMartensEtAl2019}.
A widespread area of research has explored the use of model-free reinforcement learning for spacecraft trajectory generation problems~\cite{YangHuEtAl2024,TsukamotoChungEtAl2024,MajumdarSternbergEtAl2025}.
In such approaches, the control policy is synthesized through trial-and-error in simulation without explicitly modeling the system dynamics.
However, such purely model-free methods prove brittle in the face of out-of-distribution scenarios encountered in flight and lack the flexibility to adjust constraints and specifications during mission operations.

Alternatively, a promising approach has been to formulate data-driven optimization techniques that meld machine learning and nonlinear optimization-based solution techniques for trajectory generation problems.
Such data-driven optimization approaches exploit the insight that, by solving a large number of trajectory generation problems offline, the underlying structure of the nonlinear optimization problems can be repurposed to significantly accelerate solution times onboard.
Key to such {\em amortized} optimization approaches is that they rely on using the nonlinear optimization solver onboard to ensure constraint satisfaction and certify the safety of the system, but dramatically reduce onboard solution times by learning a ``warm start'' to the optimization problem~\cite{Amos2023}.

Such amortized optimization-based approaches have been widely applied to a host of problems in aerospace applications, including for free-flying robots~\cite{BanerjeeEtAl2020}, rocket powered descent guidance~\cite{BridenJohnsonEtAl2025,BridenGurgaEtAl2025b,BridenChoiEtAl2025}, and spacecraft rendezvous and proximity operations~\cite{GuffantiGammelliEtAl2024,CelestiniAfsharradEtAl2025,TakuboGuffantiEtAl2025}.
Despite this promise, to the best of the authors' knowledge, there has yet to be an in-space demonstration of these amortized optimization-based strategies to accelerate onboard trajectory generation.
Our work builds upon~\cite{BanerjeeEtAl2020} to formulate a supervised learning-approach to warm start the~\gls{gusto} sequential convex programming solver onboard the Astrobee free-flying robot.

\section{Technical Background}~\label{sec:technical_background}

\begin{algorithm}[t]
\caption{\gls{gusto}~\cite{BonalliCauligiEtAl2019}}
\label{alg:gusto}
\begin{algorithmic}[1]
{\small
    \REQUIRE Reference trajectory $(\bar{\mathbf{x}}^0, \bar{\mathbf{u}}^0)$, Parameters $\Delta^0 > 0, \omega^\text{max} > \omega^0 \geq 1, \varepsilon > 0$; Trust region scaling $0 < \beta_\text{fail} < 1$, $\beta_\text{succ} > 1$, $0 < \rho^0 < \rho^1 < 1$, $\gamma_\text{fail} > 1$
    \FOR {$k=1,2,\dotsc,N_\text{max,iter}$}
        \STATE $(\mathbf{x}^k, \mathbf{u}^k) \gets $ solution of Eq.~\eqref{eq:ocp_cvx} about $(\mathbf{x}^{k-1}, \mathbf{u}^{k-1})$~\label{line:solve_cvx}
        \IF{$\| \mathbf{x}^{k} - \mathbf{x}^{k-1} \|_2 \leq \Delta^{k-1}$}
            \STATE Compute model accuracy ratio $\rho^{k}$~\label{line:accuracy_ratio}
            \IF {$\rho^k > \rho^1$}
                \STATE Reject solution $(\mathbf{x}^k, \mathbf{u}^k)$~\label{line:reject_soln_1}
                \STATE $\Delta^{k+1} \gets \beta_\text{fail}\Delta^k$, $\omega^{k+1} \gets \omega_k$~\label{line:trust_region_update_1}
            \ELSE
                \STATE Accept solution $(\mathbf{x}^{k}, \mathbf{u}^k)$~\label{line:accept_soln}
                \STATE $\Delta^{k+1} \gets \begin{cases}
                    \min(\beta_{\text{succ}} \Delta^k, \Delta^0) & \text{if } \rho^k < \rho^0 \\
                    \Delta^k & \text{otherwise}
                    \end{cases}$~\label{line:update_delta_k}
                \STATE $\omega^{k+1} \gets \begin{cases}
                    \omega^0 & \text{if } \bar{g}(\mathbf{x}^k, \mathbf{u}^k) \leq \varepsilon \\
                    \gamma_{\text{fail}} \omega^k & \text{otherwise}
                    \end{cases}$~\label{line:update_omega_k}
            \ENDIF
        \ELSE
            \STATE Reject solution $(\mathbf{x}^{k}, \mathbf{u}^k)$~\label{line:reject_soln_2}
            \STATE $\Delta^{k+1}\gets \Delta^k$, $\omega^{k+1} \gets \gamma_\text{fail}\omega^k$~\label{line:update_params_failure_2}
        \ENDIF
        \IF {$\omega^{k} > \omega^\text{max}$}
            \RETURN $\text{failure}$
        \ENDIF
    \ENDFOR 
    \RETURN State and control trajectories $(\mathbf{x}^k, \mathbf{u}^k)$ at iteration $k$
}
\end{algorithmic}
\end{algorithm}

In this section, we introduce the~\gls{gusto} framework~\cite{BonalliCauligiEtAl2019} for solving the trajectory generation problem and the parametric~\gls{scp} formulation we use for amortized optimization.
\subsection{Sequential Convex Programming via \gls{gusto}}
In our work, we consider parametrized trajectory optimization problems of the form,
\begin{equation} \label{eq:ocp}
\begin{array}{ll}
\underset{x_{0:N},u_{0:N}}{\textrm{minimize}} \!\!\!& \sum_{t=0}^{N} g_t(x_t,u_t;\theta) \\
\text{subject to}\!\!\!& x_0 = x_\textrm{init}(\theta),\\
& x_{t+1} = \psi_t(x_t,u_t;\theta), \hspace{0.3em} t = 0, \ldots, N-1\\
& f_{t,i}(x_t,u_t;\theta) \leq 0, \hspace{0.3em}\;\;\;\, t = 0, \ldots, N, \\
& \hspace{9em} i = 1, \ldots, n_f,
\end{array}
\end{equation}
where the state $x_t\in\mathbb{R}^{n_x}$ and control $u_t\in\mathbb{R}^{n_u}$ are continuous decision variables.
Here, the stage cost $g_t(\cdot)$ and terminal cost $g_N(\cdot)$ are assumed to be convex functions without loss of generality.
The key challenge in solving~\eqref{eq:ocp} stems from the dynamics $\psi_t (\cdot)$ and inequality constraints $f_{t,i}(\cdot)$, which are assumed smooth but non-convex.
The objective function and constraints are functions of the parameter vector $\theta \in \Theta$, where $\Theta \subseteq \mathbb{R}^{n_p}$ is the admissible set of parameters.

The~\gls{gusto} solution procedure solves Equation~\eqref{eq:ocp} in an iterative fashion by constructing a series of convex approximations,
\begin{equation} \label{eq:ocp_cvx}
\begin{array}{ll}
\underset{x_{0:N},u_{0:N}}{\textrm{minimize}} \!\!\!& \begin{aligned}
    &\sum_{t=0}^{N} g_t(x_t,u_t) + \textcolor{purple}{\omega_k \max(\hat{f}(x_t, u_t), 0)} \\
    &+ \textcolor{olive}{\omega_k \max(\| x_t - \bar{x} \|_2 - \Delta_k, 0)}
\end{aligned}\\
\text{subject to}\!\!\!& x_0 = x_\textrm{init},\\
& x_{t+1} = \hat{\psi}(x_t, u_t), \hspace{0.3em} t = 0, \ldots, N-1\\
& f_{t,i}(x_t,u_t) \leq 0, \hspace{0.3em}\;\;\;\, t = 0, \ldots, N,  i = 1,\ldots, n_f, \\
\end{array}
\end{equation}
where we have dropped the parameters $\theta$ for convenience and $\hat{\psi}(x_t, u_t)$ is the linearization of the dynamics about some reference state and control trajectories $\bar{\mathbf{x}} = (\bar{x}_0, \ldots, \bar{x}_N)$ and $\bar{\mathbf{u}} = (\bar{u}_1, \ldots, \bar{u}_N)$,
$$\hat{\psi}(x_t, u_t) = \psi_t(\bar{x}_t,\bar{u}_t) + \frac{\partial \psi_t}{\partial x}(x_t - \bar{x}_t) + \frac{\partial \psi_t}{\partial u}(u_t - \bar{u}_t),$$
and $\hat{f}(x_t, u_t)$ are the linearized constraints,
$$\hat{f}(x_t, u_t) = f_{t,i}(\bar{x}_t,\bar{u}_t) + \frac{\partial f_t}{\partial x} (x_t - \bar{x}_t) + \frac{\partial f_t}{\partial u} (u_t - \bar{u}_t),$$
We note that, in accordance with the~\gls{gusto} solution procedure, $\hat{f}(\cdot)$ is included in penalty form using the $\textcolor{purple}{\omega_k\max(\hat{f}(x_t, u_t), 0)}$ operation, which is a non-smooth expression that can be rewritten using linear constraints~\cite{NocedalWright2006}.
Finally, a trust region constraint $\| x_t - \bar{x}_t\| \leq \Delta_k$ facilitates improved convergence by restricting the solution of the convex approximation to remain ``close'' to the linearization $(\bar{x}, \bar{u})$ and this constraint is similarly rewritten in penalty form as $\textcolor{olive}{\omega_k\max(\| x_t - \bar{x} \|_2 - \Delta_k, 0)}$.

The full~\gls{gusto} solution procedure for solving the non-convex trajectory generation problem is given in Algorithm~\ref{alg:gusto}
The parameters $\Delta^0$ and $\omega^0$ initialize the trust region constraint and the relative cost weightings, respectively.
\gls{gusto} commences with an initial reference trajectory $(\bar{\mathbf{x}}^0, \bar{\mathbf{u}}^0)$, where $\bar{\mathbf{x}}^0 = (\bar{x}_0^0, \ldots, \bar{x}_N^0)$ and $\bar{\mathbf{u}}^0 = (\bar{u}_1^0, \ldots, \bar{u}_N^0)$.
This reference trajectory is used to construct the convex problem given in Eq.~\eqref{eq:ocp_cvx} and recover the new solution $(\mathbf{x}^k, \mathbf{u}^k)$ (Line~\ref{line:solve_cvx}).
If the new solution $(\mathbf{x}^k, \mathbf{u}^k)$ violates the trust region constraint, then the solution is rejected (Lines~\ref{line:reject_soln_2}-\ref{line:update_params_failure_2}).
Otherwise, an additional check proceeds by computing the model accuracy ratio $\rho^k$ defined in Equation 5 from~\cite{BonalliCauligiEtAl2019} (Line~\ref{line:accuracy_ratio}).
If $\rho^k > \rho^1$, then the new solution is rejected (Lines~\ref{line:reject_soln_1}-\ref{line:trust_region_update_1}).
Otherwise, the solution is accepted (Line~\ref{line:accept_soln}) and the parameters $\Delta^k$ and $\omega^k$ also updated (Lines~\ref{line:update_delta_k}-\ref{line:update_omega_k}).

\begin{algorithm}[t]
\caption{Learning the Problem-Solution Mapping}
\label{alg:learning_offline}
\begin{algorithmic}[1]
{\small
    \REQUIRE Training parameters $\{ \theta_i \}_{i=1}^{N_d}$, batch size $N_\text{bs} < N_d$ and $N_\text{epoch}$ training epochs.~\label{line:sample_params}
    \STATE Initialize training batch $\mathcal{D} \gets \emptyset$ and randomized neural network parameters $\phi^0$.
    \FOR {$i = 1, \ldots, N_d$}
        \STATE $(\mathbf{x}_i^*, \mathbf{u}_i^*) \gets \text{GuSTO}(\theta_i)$~\label{line:solve_with_gusto}
        \STATE $\mathcal{D} = \mathcal{D} \cup \{ \theta_i, (\mathbf{x}_i^*, \mathbf{u}_i^*) \}$~\label{line:add_to_training_set}
    \ENDFOR
    \FOR {$k=1,\ldots,N_\text{epoch}$}
        \STATE Sample batch $\{ \theta_i, (\mathbf{x}_i^*, \mathbf{u}_i^*) \}_{i=1}^{N_\text{bs}}$ from $\mathcal{D}$~\label{line:sample_training_data}
        \STATE $\phi^k \gets \text{SGD}(\phi^{k-1}, \{ \theta_i, (\mathbf{x}_i^*, \mathbf{u}_i^*) \}_{i=1}^{N_{bs}})$~\label{line:sgd_update}
    \ENDFOR
    \RETURN Trained neural network parameters $\phi^k$
}
\end{algorithmic}
\end{algorithm}

\subsection{Learning the Optimal Control Structure}
The key insight of amortized optimization is that the solution to the optimization problem in Eq.~\eqref{eq:ocp} is largely influenced by the parameters $\theta \in \Theta$ and that there exists a ``problem-solution'' mapping between parameters $\theta$ and the solution $(\mathbf{x}^*, \mathbf{u}^*)$.
As such, the goal of amortized optimization is to learn this mapping in a data-driven fashion by simulating various parameters $\theta$ from representative problems of interest and repurposing this learned mapping to accelerate solution times onboard for new problems.

Algorithm~\ref{alg:learning_offline} sketches the high-level overview of the training procedure used to learn this problem-solution mapping for trajectory generation problems as found in~\cite{BanerjeeEtAl2020,CauligiCulbertsonEtAl2022}.
The procedure begins by sampling representative parameters for problems of interest, \eg{}, initial and goal conditions, obstacles, among others (Line~\ref{line:sample_params}).
Next, the optimal solution $(\mathbf{x}_i^*, \mathbf{u}_i^*)$ is solved for using~\gls{gusto} and added to the training set (Lines~\ref{line:solve_with_gusto}-\ref{line:add_to_training_set}).
Finally, the neural network parameters $\phi$ are updated using stochastic gradient descent to minimize the loss function, \ie{}, cross-entropy loss for classification and mean squared error for regression formulations (Lines~\ref{line:sample_training_data}-\ref{line:sgd_update}).

\section{Approach}
We utilize a two-stage approach for learning the structure of the Astrobee motion planning problem, following the framework of~\cite{BanerjeeEtAl2020}.
In the first stage, we generate training data offline using a simple simulator of the~\gls{iss} presented in~\cite{BualatSmithEtAl2018} and use the GuSTO algorithm~\cite{BonalliCauligiEtAl2019,BonalliBylardEtAl2019} to find a local solution for the Astrobee motion planning problem.
We then train a neural network to learn the problem-solution mapping between the motion planning problem parameters to the solution found by GuSTO.
In the second stage, this learned neural network is deployed online and used to generate candidate warm start solutions given new motion planning queries.
This warm start is then provided to the GuSTO solver to enforce runtime safety constraints for the system. This approach is illustrated in Figure~\ref{fig:flowchart}.

\subsection{Astrobee Optimal Control Formulation}
Here, we formulate the optimal control formulation used for the Astrobee free-flyer trajectory generation problem.

{\em System Dynamics: }
The trajectory generation problem for Astrobee entails solving for a six degree-of-freedom state trajectory and a control trajectory including translational and rotational inputs. 
Specifically, the 13 dimensional state for the free-flying spacecraft robot model consists of position $r\in\mathbb{R}^3$, velocity $v\in\mathbb{R}^3$, the quaternion representation of attitude $q \in \mathcal{S} \subseteq \mathbb{R}^4$, and angular velocity $\omega\in\mathbb{R}^3$, and the control variables are the force $F\in\mathbb{R}^3$ and moment $M\in\mathbb{R}^3$.
The continuous-time dynamics are given by
\begin{align}
\begin{pmatrix} \dot{r}_t\\ \dot{v}_t\\ \dot{q}_t\\ \dot{\omega}_t \end{pmatrix} =
\begin{pmatrix}
v_t\\ F_t / m \\ \frac{1}{2} \Xi (\omega_t) q_t \\
J^{-1}(M_t-\omega_t\times J \omega_t)
\end{pmatrix},~\label{eq:six_dof_dyn}
\end{align}
where $m$ and $J$ are the robot mass and inertia tensor, respectively, and $\Xi (\omega)$ is the quaternion kinematics matrix~\cite{Shuster1993}.
State constraints for this system include norm bounds for velocity and angular velocity, as well as norm bound control constraints for the force and moment.
We further enforce a non-convex equality constraint to satisfy the quaternion manifold constraint,
\begin{align}
\| q_t\|_2 = 1.~\label{eq:quat_norm_eq}
\end{align}

{\em Boundary Conditions: }
\begin{align}
x_0 = x_\text{init}, \quad x_N \in \mathcal{X}_\text{goal},~\label{eq:bc}
\end{align}
where $x_\text{init}$ is the initial state provided by state estimation and $\mathcal{X}_\text{goal}$ is the terminal set defined within a tolerance $\delta_\text{goal} > 0$ about the goal position $r_\text{goal} \in \mathbb{R}^3$,
\begin{align*}
\mathcal{X}_\text{goal} = \{x_t \quad | \|r_t - r_\text{goal}\|_2 \leq \delta_\text{tol} \}.
\end{align*}

{\em Vehicle Constraints: }
To satisfy Astrobee vehicle limits, we enforce conic bounds on the speeds,
\begin{align}
\| v_t \|_2 \leq v_\text{max},~\label{eq:vel_bound}\\
\| \omega_t \|_2 \leq \omega_\text{max},~\label{eq:om_bound}
\end{align}
where $v_\text{max}$ and $\omega_\text{max}$ are the maximum translational and rotational speeds, respectively.
Finally, we enforce actuator limits on the forces and moments,
\begin{align}
\| F_t \|_2 \leq F_\text{max},~\label{eq:force_con}\\
\| M_t \|_2 \leq M_\text{max}.~\label{eq:moment_con}
\end{align}

{\em Collision Avoidance Constraints: }
For collision avoidance with surrounding obstacles, we enforce a safety constraint using the signed distance function~\cite{SchulmanDuanEtAl2014},
\begin{align}
\text{sd}(x_t) > \delta_\text{sd},~\label{eq:obs_avoid}
\end{align}
where a negative value $\text{sd}(x_t) < 0$ indicates collision with an obstacle and we seek to enforce a minimum clearance $\delta_\text{sd} > 0$.
For this work, we assume axis-aligned bounding boxes only and compute the signed distance function manually by assuming a simple spherical robot footprint for Astrobee.

{\em Trajectory Generation Problem: }
Collecting these constraints, the nonlinear optimization problem used to model the Astrobee trajectory generation problem is given by,
\begin{align}
    \underset{\mathbf{x}, \mathbf{u}}{\textrm{minimize}} \,\, & \sum_{t=0}^{N} u_t^T R u_t,~\label{eq:traj_gen}\\
    \textrm{subject to} 
    \,\,& \eqref{eq:six_dof_dyn}, \eqref{eq:quat_norm_eq},\eqref{eq:bc}, \eqref{eq:vel_bound}, \eqref{eq:om_bound}, \eqref{eq:force_con}, \eqref{eq:moment_con}, \eqref{eq:obs_avoid}. \nonumber
\end{align}
Here, the non-convex constraints include the rotational kinematics and dynamics (Eq.~\eqref{eq:six_dof_dyn}), the quaternion norm constraint (Eq.~\eqref{eq:quat_norm_eq}), and the collision avoidance constraint (Eq.~\eqref{eq:obs_avoid}). This optimization problem can be solved using the GuSTO solver in Algorithm~\ref{alg:gusto}.

\subsection{Supervised Learning from Optimized Trajectories}

We train a neural network to predict trajectory solutions for new planning problems by learning from optimized examples. Training data is generated offline by sampling representative problem parameters $\theta$, which include the robot's start and goal states as well as virtual obstacles.

We sample problems from two simulation environments provided by the Astrobee software suite: (i) a planar $2\,\text{m} \times 2\,\text{m}$ granite table simulating the testbed at NASA Ames, and (ii) the Japanese Experiment Module (JEM) on the~\gls{iss}, approximated as a $1.5\,\text{m} \times 6.4\,\text{m} \times 1.7\,\text{m}$ 3D volume in~\gls{iss} coordinates. Start and goal positions are sampled uniformly in each environment, with start and goal attitudes sampled uniformly over $\mathrm{SO}(3)$. All trajectories begin and end with zero linear and angular velocity. Obstacles are sampled as axis-aligned cuboids with random dimensions and are placed within $1\,\text{m}$ of the workspace origin.

For each sampled problem, we formulate a trajectory optimization problem as defined in Eq.~\eqref{eq:traj_gen}, incorporating dynamics and collision-avoidance constraints. We solve each instance using the GuSTO algorithm (Alg.~\ref{alg:gusto}), producing optimized state and control trajectories $(\mathbf{x}, \mathbf{u})$.

To prepare the data for learning, each dimension of the state and control trajectories is fit with a $p^{\text{th}}$-order polynomial in time, where $p = 3$. At each timestep $t$, the parameterizations are:
\begin{align}
\mathbf{x}_t &= \sum_{j=0}^p \boldsymbol{\alpha}_j t^j, \label{eq:state_param}\\
\mathbf{u}_t &= \sum_{j=0}^p \boldsymbol{\beta}_j t^j \label{eq:control_param}
\end{align}
where $\boldsymbol{\alpha}_j$ and $\boldsymbol{\beta}_j$ are the polynomial coefficients. The state is 13-dimensional and the control is 6-dimensional, resulting in an output size of $(13 + 6) \times 4 = 76$.

The neural network receives as input the problem specification: the start and goal poses (7D each, consisting of position and quaternion) and a fixed-size 6D vector representing the obstacle geometry, for a total of 20 input dimensions. It outputs the fitted polynomial coefficients $(\boldsymbol{\alpha}_j, \boldsymbol{\beta}_j)$. The network is a fully connected feedforward model with three hidden layers of sizes 256, 512, and 256, with ReLU activations and uniformly initialized weights. Training minimizes the mean squared error between the predicted and ground truth coefficients.

The final training set consists of approximately 11,000 trajectories sampled from the~\gls{iss} environment and 2,000 trajectories from the granite table. Details of the simulation environments are provided in the next section.

\section{Experimental Setup}

\begin{figure}[tbh]
    \centering
    \begin{subfigure}[t]{0.24\textwidth}
        \centering
        \includegraphics[width=\linewidth]{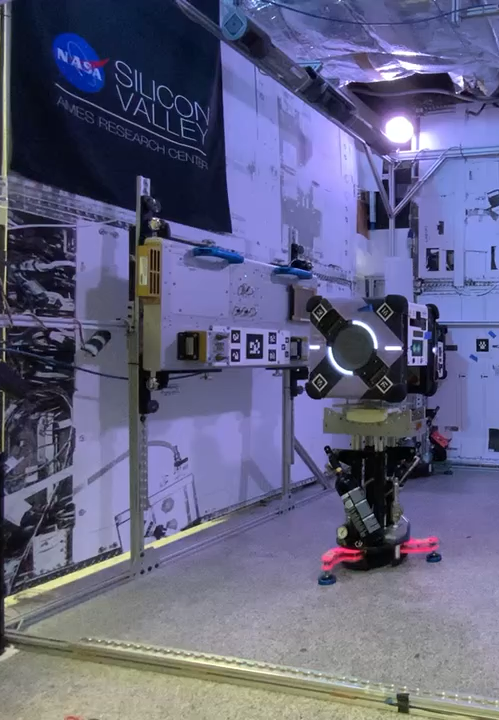}
        \caption{Ground testing at NASA Ames}
        \label{fig:testing_granite}
    \end{subfigure}%
    \hfill
    \begin{subfigure}[t]{0.24\textwidth}
        \centering
        \includegraphics[width=\linewidth]{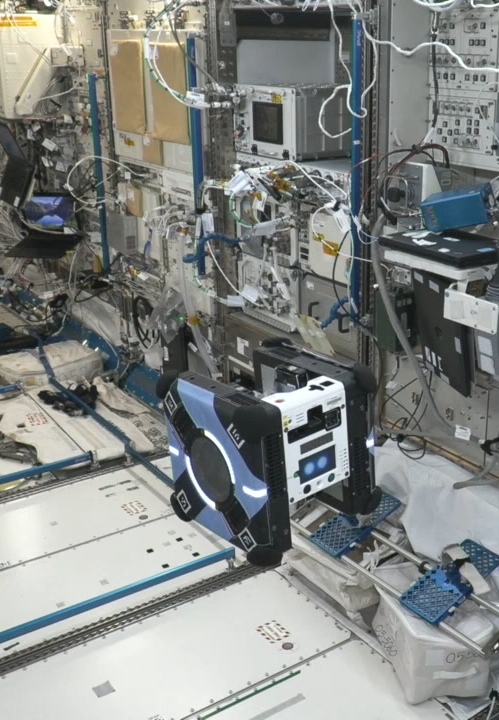}
        \caption{ISS testing}
        \label{fig:testing_iss}
    \end{subfigure}
    \caption{\textbf{Testing phases for software on Astrobee.} Left: Ground testing on a 2D air-bearing granite table at NASA Ames to validate software on real hardware in a controlled planar environment. Right: On-orbit flight testing aboard the ISS, where Astrobee executed autonomous trajectories in a 6-DOF microgravity environment.}
    \label{fig:testing_pics}
\end{figure}

The Astrobee platform is a free-flying, 30-cm wide, cube-shaped robot developed by NASA Ames Research Center to enable a wide range of autonomous manipulation and inspection tasks aboard the~\gls{iss}~\cite{BualatSmithEtAl2018}.
Astrobee is a holonomic robot with six degrees-of-freedom, actuated by twelve independent thrusters that draw air from the~\gls{iss} cabin environment.

This section describes the key software components, testing environments, and unique challenges involved in integrating and validating our trajectory optimization software on the Astrobee platform across simulation, ground, and flight experiments.

\subsection{Software Integration}
Our software was implemented in C\texttt{++} and integrated into a fork of NASA's Astrobee flight software (FSW)~\cite{Astrobee, BualatSmithEtAl2018, SmithBarlowEtAl2016}, which is publicly available\footnote{\url{https://github.com/StanfordASL/astrobee/tree/scp_traj_opt}}.
Astrobee FSW runs on three onboard processors: an ARM-based low-level processor (LLP) for propulsion control, another ARM-based mid-level processor (MLP) for the primary flight software, and an Android-based high-level processor (HLP) for relaying and processing guest science commands~\cite{fluckiger2018astrobee}. Our planner executed on the MLP, with lightweight modifications to the HLP interface for experiment coordination.

Astrobee's FSW uses the Robot Operating System (ROS) middleware, with approximately 46 nodelets grouped into 14 processes. The default framework consists of planning (Planner), mapping (Mapper), localization (Extended Kalman Filter EKF), control (CTL), and a force allocation module (FAM), all orchestrated by a managing node, the choreographer~\cite{fluckiger2018astrobee}. Our trajectory optimization was implemented as a new planner node, `planner\_scp', that subscribes and publishes to the same topics as the existing planners, namely `planner\_qp' and `planner\_trapezoidal', based on quadratic programming (QP) and trapezoidal motion planning respectively. The choreographer node was configured to call our planner as an additional option.

External libraries were integrated within the Astrobee FSW stack, specifically OSQP~\cite{StellatoBanjacEtAl2020} for solving the optimization problem and LibTorch~\cite{PaszkeGrossEtAl2017} for machine learning. Cross-compilation of these libraries, particularly LibTorch, was a significant challenge, as discussed later in this section. However, inference of the lightweight feedforward model was sufficiently performant on the MLP.

\subsection{Testing Environments}
Our testing followed a three-phase progression: simulation, ground testing, and finally, on-orbit flight testing. 

\paragraph{Simulation}
The Astrobee software stack includes a high-fidelity Gazebo-based simulator with two environments: a planar 3-DOF ``granite table'' world that emulates the facility at NASA Ames, and a 6-DOF~\gls{iss} world based on the Japanese Experiment Module (JEM) within which Astrobee operates. Simulation allowed early software validation in both 2D and 3D microgravity scenarios.

\paragraph{Ground Testing}
The integrated software was tested on Astrobee hardware at the NASA Ames granite table facility, where Astrobee floats on a 2D air-bearing platform within a mock-up of the~\gls{iss} interior, as shown in Figure~\ref{fig:testing_granite}. Over multiple day-long test sessions, we iterated on algorithm tuning and operations design, with support from the Ames team.

\paragraph{Flight Testing}
Flight experiments were performed aboard the~\gls{iss} as a ``crew-minimal'' operation, involving 35 minutes of crew time  for setup and teardown, and four hours of autonomous operation. The concept of operations (CONOPS) involved undocking Astrobee, moving it to a designated ``HOME'' pose, and executing multiple round-trip trajectories beginning and ending at ``HOME''. Each trajectory was executed twice: once from a cold start and once from a warm start that required neural network inference onboard.

Commands for trajectory start and goal poses, warm/cold initialization settings, and virtual obstacle placement were sent by ground operators, who remained on standby for teleoperation if needed. A total of 18 trajectories (each with multiple sub-segments) were executed during the test session. A second Astrobee, configured identically, remained docked as a backup but was not needed. After the flight, experiment logs, images, and videos were downlinked for post-processing and analysis (the results of which are presented in Section~\ref{sec:results}). A short video is available on YouTube\footnote{\url{https://youtu.be/NVDMd88XFhs}}. 

\subsection{Challenges}
Cross-compilation of external libraries posed a significant challenge for hardware integration, as was also noted in prior work on Astrobee hardware~\cite{doerr2024reswarm}. Nonetheless, this effort achieved the first successful cross-compilation of PyTorch as a learning-based control framework on Astrobee.

A practical hurdle was tuning the runtime of the planner to operate within the timing constraints of Astrobee’s flight software, particularly the choreographer node. Processor load varied across trajectories and testing sessions, affecting how many optimization iterations could be completed in time. We addressed this by introducing dynamic ROS parameters to adjust iteration limits, convergence tolerances, and time discretization, to achieve reliable and robust planner performance.

Another challenge was the performance of Astrobee's localization module, which relies on detecting visual features in a known map of the~\gls{iss}. Some areas of the JEM module are feature-sparse, leading to degraded pose estimates and drift, even when the planner produced optimal trajectories. To mitigate this, we worked with the NASA Ames team to identify feature-rich regions, minimized rotational motion to improve tracking, and implemented abort logic with teleop fallback to recover from localization failures.

\section{Results and Discussion}
\label{sec:results}

\begin{table*}[b]
    \centering
    \caption{\textbf{Comparison of cold vs. warm start for average optimal cost and iterations / run time required to converge.} Columns are trajectory categories: translation only, translation+rotation, seen (in-distribution) obstacles, and unseen (out-of-distribution) obstacles.}

    \renewcommand{\arraystretch}{1.2}
    \begin{tabular}{l|cccc|cccc}
        \toprule
        \multirow{2}{*}{\textbf{Method}} 
        & \multicolumn{4}{c|}{\textbf{Optimal cost $\downarrow$}} 
        & \multicolumn{4}{c}{\textbf{\# Iterations to converge (run time in seconds) $\downarrow$}} \\
        \cmidrule(lr){2-5} \cmidrule(lr){6-9}
        & \textbf{TransOnly} & \textbf{Trans+RotOnly} & \textbf{Obs\_Seen} & \textbf{Obs\_OOD} 
        & \textbf{TransOnly} & \textbf{Trans+RotOnly} & \textbf{Obs\_Seen} & \textbf{Obs\_OOD} \\
        \midrule
        \textbf{SCP cold}       & \textbf{0.6311} & \textbf{0.0025} & 0.5700 & \textbf{0.5844}  & \textbf{2096 (8.3\,s)} & 634 (2.6\,s) & 4182 (14.3\,s) & \textbf{2855 (11.2\,s)} \\
        \textbf{SCP warm}       & \textbf{0.6326} & \textbf{0.0024} & \textbf{0.3902} & 0.6529  & 2357 (8.9\,s) & \textbf{254 (1.1\,s)} & \textbf{2106 (7.0\,s)} & 3065 (12.7\,s) \\
        \bottomrule
    \end{tabular}
    \label{tab:method_comparison}
\end{table*}

\begin{figure*}[bt]
    \centering
    \begin{subfigure}[t]{0.32\textwidth}
        \centering
        \includegraphics[width=\linewidth]{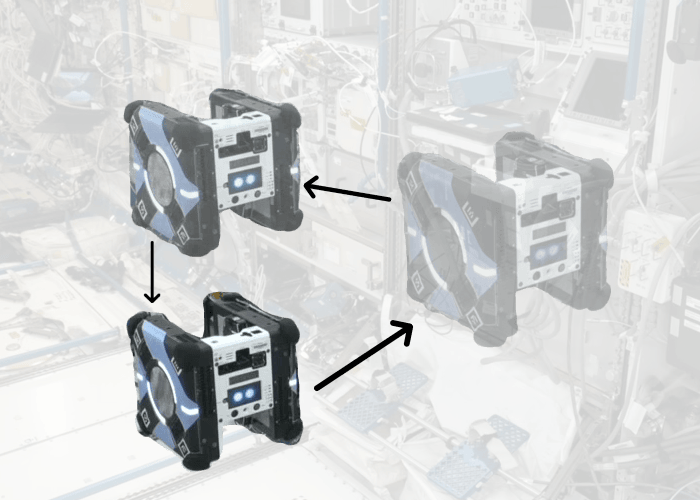}
        \phantomcaption
        \label{fig:dummy1}
    \end{subfigure}%
    \hfill
    \begin{subfigure}[t]{0.32\textwidth}
        \centering
        \includegraphics[width=\linewidth]{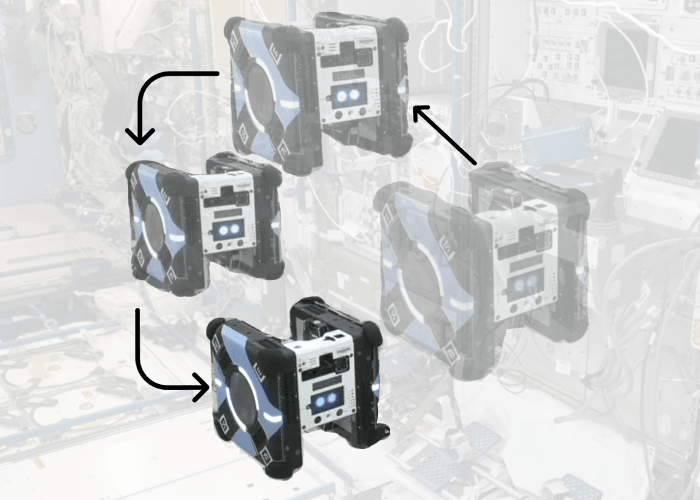}
        \phantomcaption
        \label{fig:dummy2}
    \end{subfigure}%
    \hfill
    \begin{subfigure}[t]{0.32\textwidth}
        \centering
        \includegraphics[width=\linewidth]{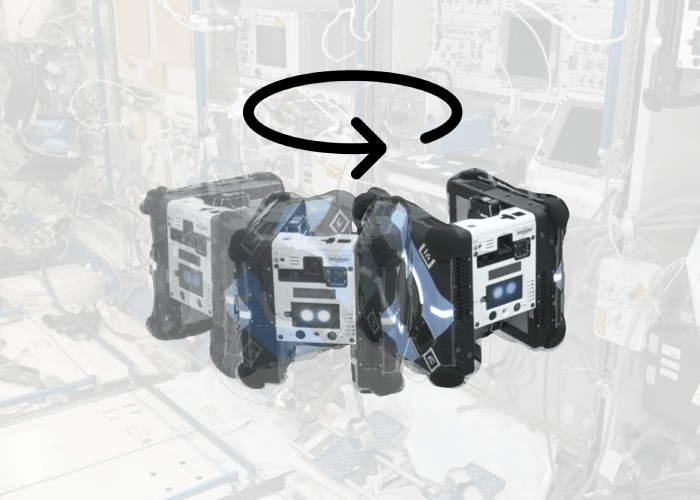}
        \phantomcaption
        \label{fig:dummy3}
    \end{subfigure}
    
    \begin{subfigure}[t]{0.32\textwidth}
        \centering
        \includegraphics[width=\linewidth, trim=0 0 0 40, clip]{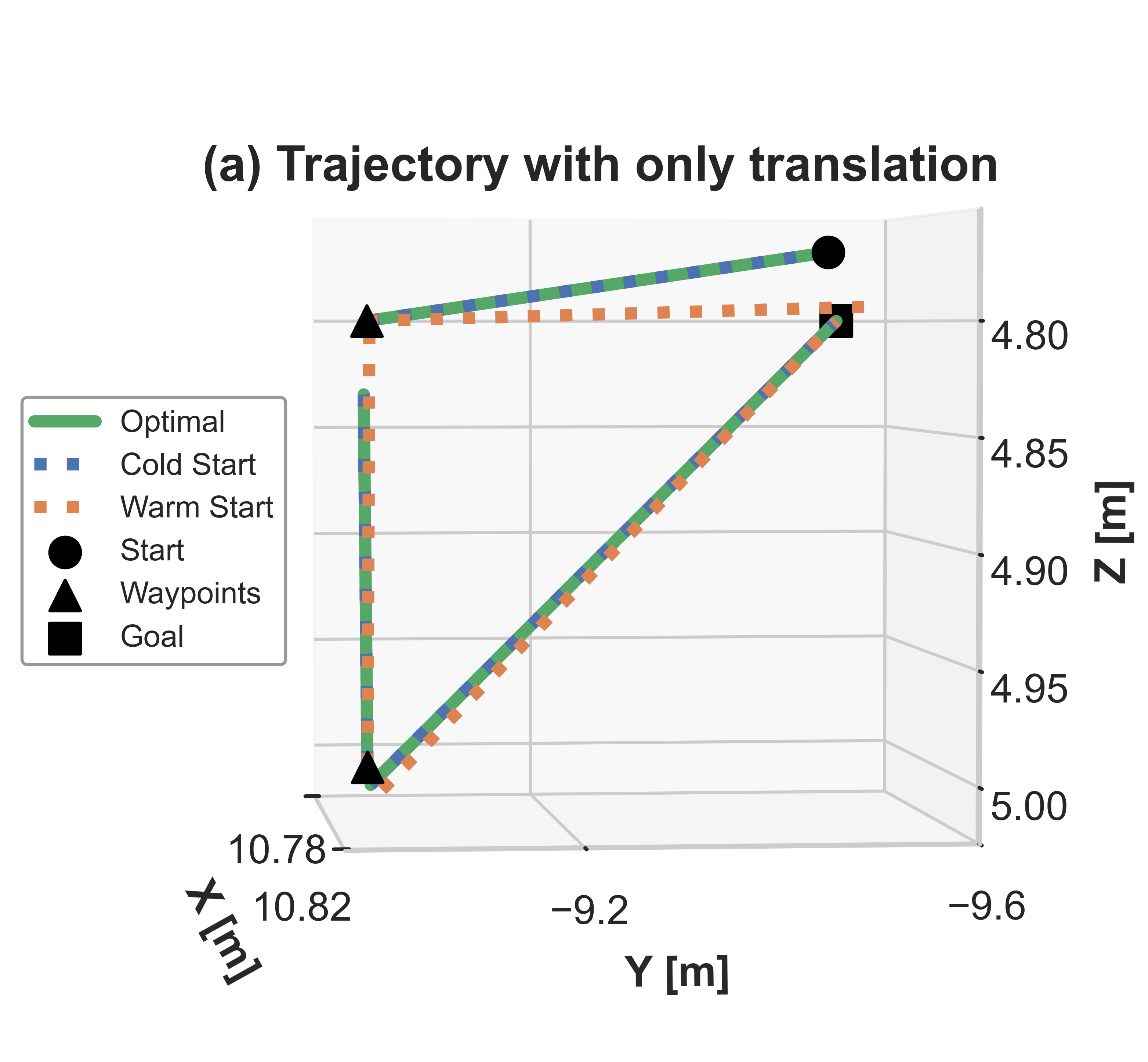}
        \phantomcaption
        \label{fig:traj_translation}
    \end{subfigure}%
    \hfill
    \begin{subfigure}[t]{0.32\textwidth}
        \centering
        \includegraphics[width=\linewidth, trim=0 0 0 40, clip]{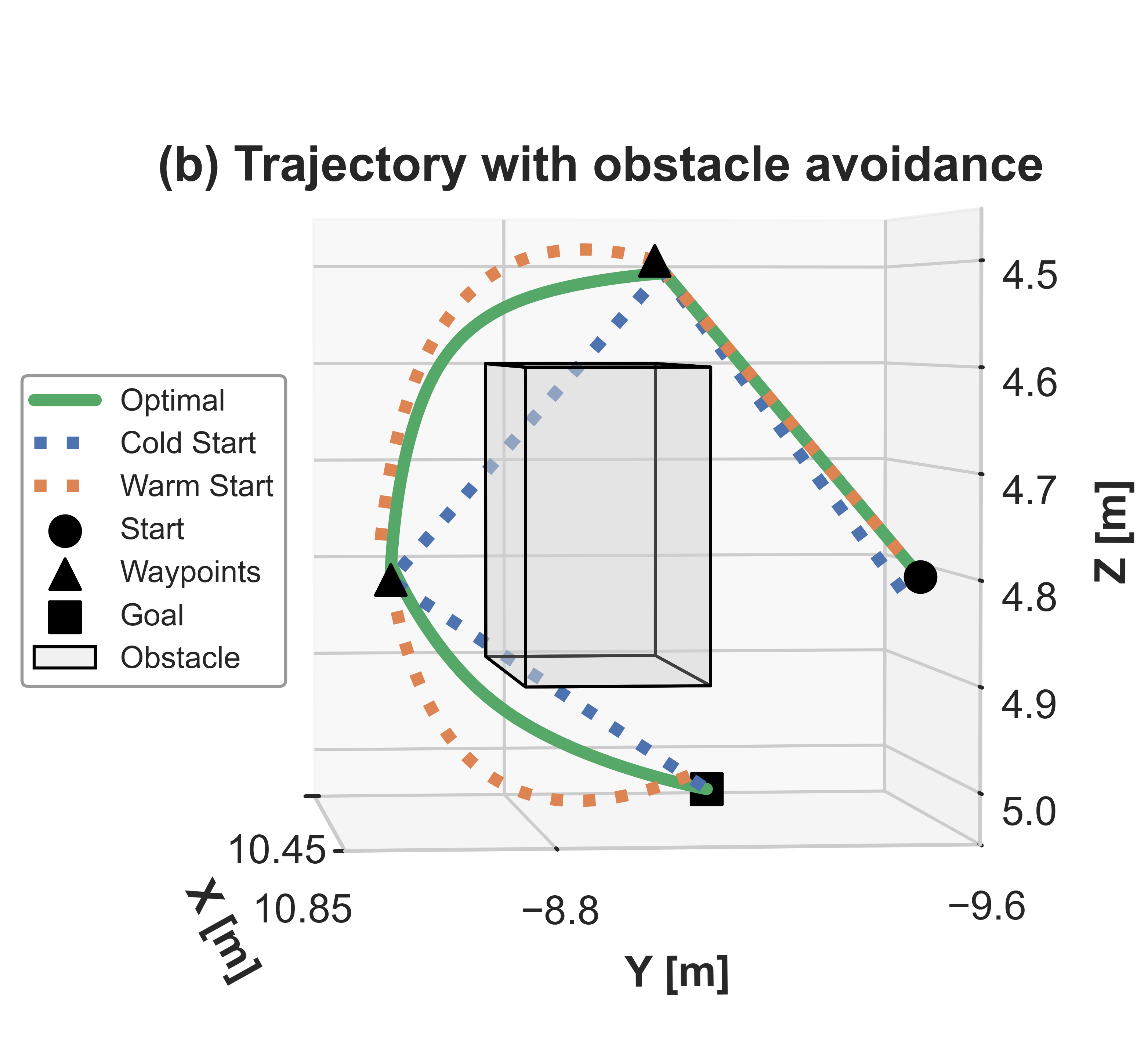}
        \phantomcaption
        \label{fig:traj_obs_avoidance}
    \end{subfigure}%
    \hfill
    \begin{subfigure}[t]{0.32\textwidth}
        \centering
        \includegraphics[width=\linewidth, trim=0 0 0 0, clip]{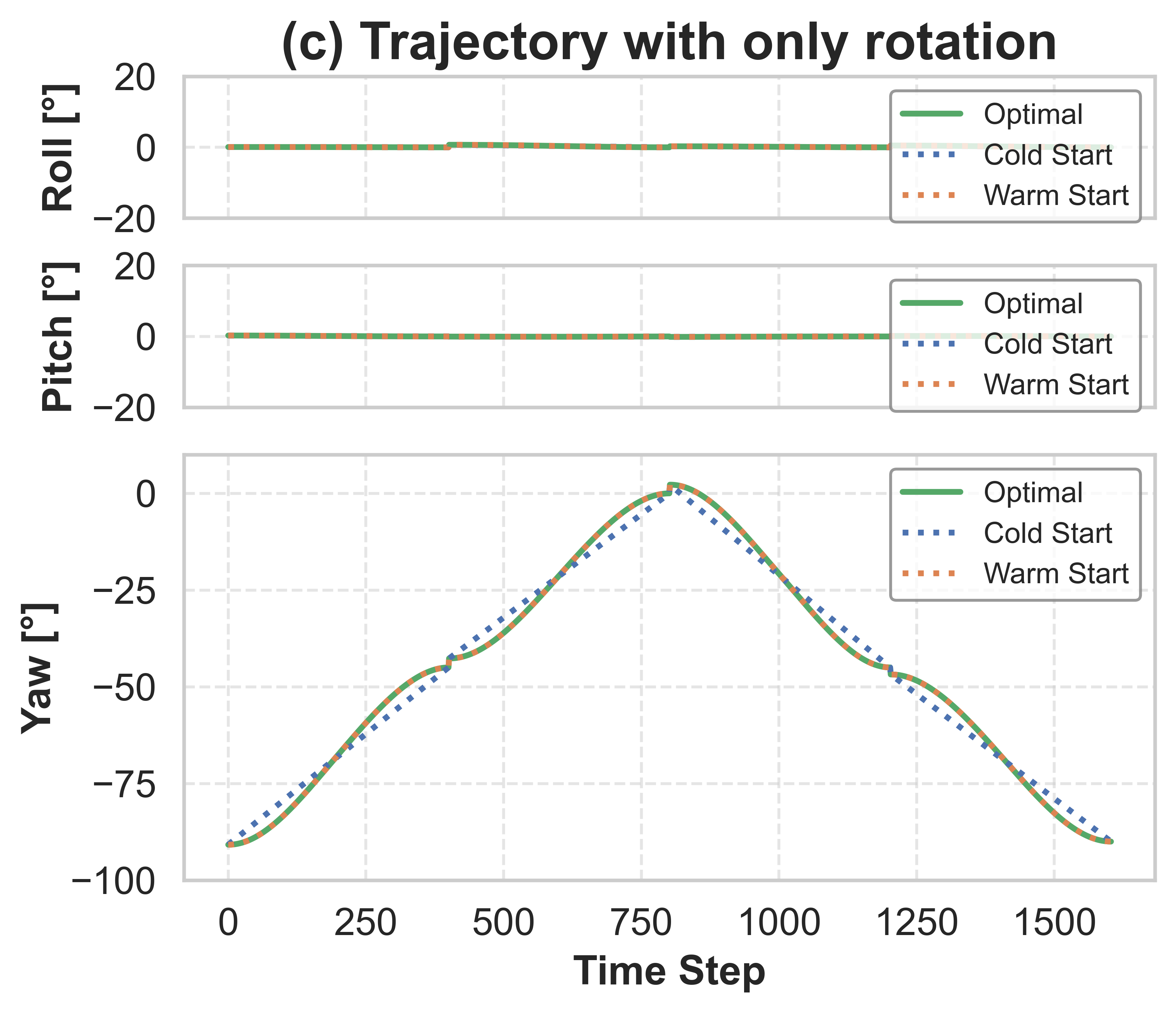}
        \phantomcaption
        \label{fig:traj_rotation}
    \end{subfigure}
    \vspace{-1em}
    \caption{\textbf{Examples of trajectories tested.} Top row shows Astrobee execution and bottom row shows the plotted data. Both \textcolor[HTML]{4C72B0}{{cold}} and \textcolor[HTML]{DD8452}{{warm}} starts result in the same \textcolor[HTML]{55A868}{optimal} trajectory. Warm starts are qualitatively closer to the optimal trajectory, especially for trajectories with obstacle avoidance and rotations.}
    \label{fig:traj_examples}
\end{figure*}

\textbf{Learned Warm Starts Accelerate Convergence Without Sacrificing Optimality:} 
A core hypothesis of this work is that learned warm starts can accelerate trajectory optimization without degrading solution quality. Table~\ref{tab:method_comparison} confirms that warm and cold starts achieve nearly identical optimal costs across all trajectory categories, including highly non-convex cases with rotation and obstacle avoidance. Figure~\ref{fig:traj_examples} illustrates that both approaches converge to the same trajectory, with warm starts beginning closer to the solution, reflecting the robustness of SCP methods~\cite{MaoSzmukEtAl2016}.  

\begin{figure}[tbh]
  \centering
  \includegraphics[width=0.38\textwidth, trim=0 10 0 20, clip]{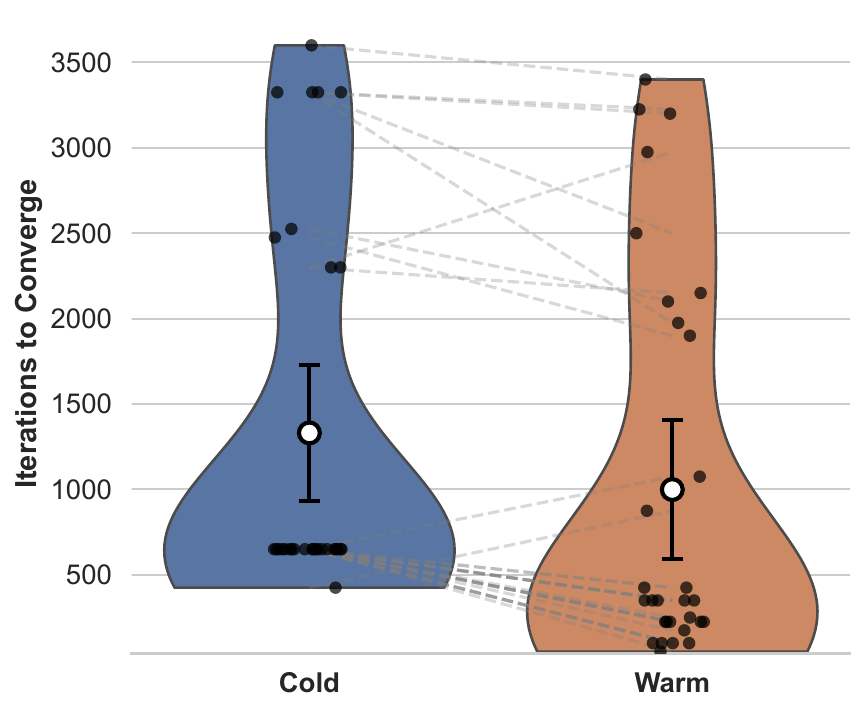}
  \caption{\textbf{Comparison of convergence iterations for cold and warm starts}. Each pair of connected points represents a matched run, i.e., a \textcolor[HTML]{4C72B0}{{cold}} and \textcolor[HTML]{DD8452}{{warm}} start for the same start and goal states and obstacle constraints. The plot shows a trend towards reduction in iterations required for convergence when using a warm start. Black markers indicate group means, with vertical bars showing 95\% confidence intervals computed using the standard error of the mean.}
  \label{fig:cold_vs_warm}
\end{figure}

The primary benefit of warm starts lies in convergence speed. Figure~\ref{fig:cold_vs_warm} shows per-instance reductions in solver iterations: each line links a cold and warm start for the same problem instance, black markers denote sample means, and bars show 95\% confidence intervals. The predominance of downward-sloping lines demonstrates statistically significant acceleration across problem instances. The few upward-sloping outliers arise from out-of-distribution obstacle scenarios, discussed as a limitation later. This confirms that warm-starting is an effective and practical strategy for real-time trajectory optimization.



\textbf{Iteration Gains Are Most Pronounced in Non-Convex Regimes:} 
To analyze when warm starting is helpful, we partition the trajectories into four categories: (1) translation only, (2) rotation only, (3) obstacle avoidance with seen (in-distribution) obstacles, and (4) obstacle avoidance with unseen (out-of-distribution) obstacles.

\begin{figure}[t]
  \centering
  \includegraphics[width=0.49\textwidth, trim=0 0 0 0, clip]{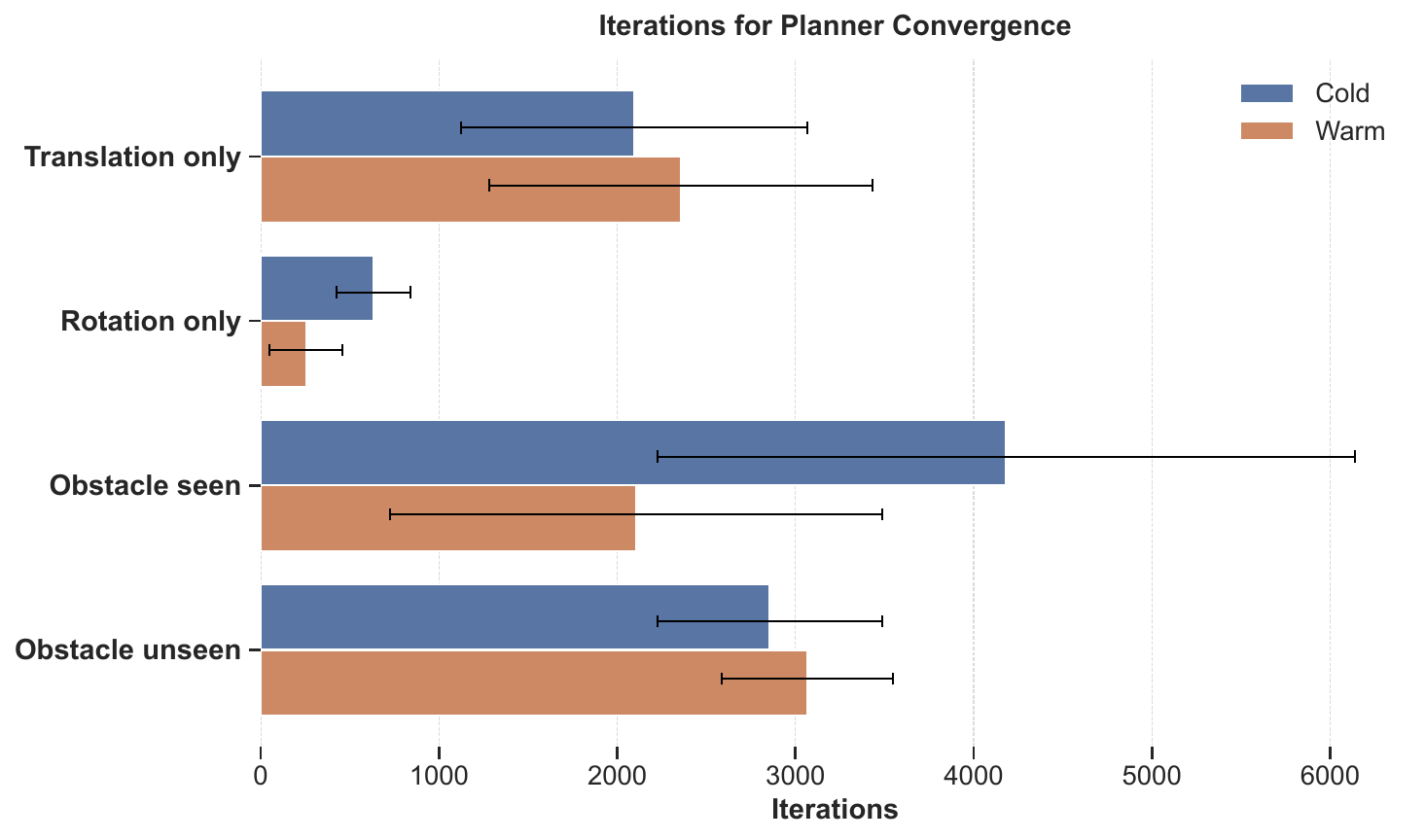}
  \caption{\textbf{Iterations to convergence for cold and warm starts across trajectory categories}. Horizontal bars indicate mean iterations required for planner convergence under \textcolor[HTML]{4C72B0}{{cold}} and \textcolor[HTML]{DD8452}{{warm}} start conditions, with error bars showing one standard deviation. Warm starts perform similar to cold starts for translation-only trajectories. Warm starts require fewer iterations than cold starts for trajectories with rotation and seen obstacles,~\ie{}, non-convexities learned by the model. However, the model did not generalize to unseen obstacles.}
  \label{fig:planner_iterations}
\end{figure}

For translation-only trajectories, Figure~\ref{fig:planner_iterations} shows that warm and cold starts converge in nearly the same number of iterations. The problem is convex, so learned priors are unnecessary. In these convex translation-only cases, the relatively large iteration count ($\approx 2000$) arises from the numerical settings rather than problem difficulty. Each trajectory was discretized at 20~Hz over long horizons (up to 40~s), yielding hundreds of timesteps ($N=800$) and tens of thousands of decision variables. We also enforced tight solver tolerances ($10^{-5}$--$10^{-8}$) to ensure that the fine-grained trajectories remained dynamically consistent with Astrobee’s onboard controller, with strict tracking tolerances. This high-accuracy configuration improves hardware robustness but increases the conditioning of the quadratic program and, consequently, the number of ADMM iterations required by OSQP. In contrast, our earlier work~\cite{BanerjeeEtAl2020} solved the same trajectory optimization problems with coarser discretizations ($N \le 100$) and looser tolerances, converging in only 10--100 iterations.

In contrast to the lack of improvement in translation-only cases, the rotation-only trajectories show a significant benefit,~\ie{}, warm starts reduce iteration counts by over 60\% on average. This is because attitude dynamics, parameterized by quaternions, are non-convex, and poor initializations can violate feasibility or cause slow convergence. The learned model implicitly encodes rotational structure from prior data, enabling faster convergence by proposing trajectories that satisfy rotational feasibility earlier in the optimization. This ability of the warm start to encode the rotational structure is qualitatively visible in Figure~\ref{fig:traj_rotation}.

Obstacle avoidance tasks also show a large acceleration,~\ie{}, up to 50\% fewer iterations when obstacles match those seen in the training distribution.  The warm start model implicitly learns common avoidance patterns and initializes trajectories that navigate around the obstacle effectively. The SCP solver benefits from this informed guess, requiring fewer iterations to converge to a feasible and optimal trajectory. However, this benefit vanishes for unseen obstacles.

\textbf{Generalization to Unseen Obstacles is Limited by Model Architecture:} 
Performance degrades when obstacle configurations deviate from the training distribution. For obstacles not seen during training (Figure~\ref{fig:planner_iterations}, bottom row), warm starts provide no benefit and exhibit higher variance. In most of these cases, warm starts initialize with infeasible trajectories,~\eg{}, passing through the obstacle, which the optimizer must then repair over many iterations. The iteration counts in these cases match or exceed those of cold starts.

This result reflects the limited expressivity of the current feedforward architecture, which lacks spatial reasoning and contextual understanding. As shown in prior work~\cite{shi2022neural, li2022trajectory}, feedforward models struggle to generalize when training coverage is insufficient. More expressive architectures,~\eg{}, transformer-based priors with attention mechanisms~\cite{BridenGurgaEtAl2025b, GuffantiGammelliEtAl2024}, could better capture task context and improve generalization in unfamiliar environments. Another approach to mitigate unseen obstacles in low-dimensional systems is to use an RRT planner to generate a feasible initial trajectory. 

\textbf{Trajectory Visualizations Reveal Qualitative Differences in Initialization:} 
Figure~\ref{fig:traj_examples} shows example trajectories across categories. In translation-only tasks, both warm and cold trajectories begin reasonably close to the solution, converging quickly. Obstacle avoidance tasks illustrate how warm starts tend to choose a path that matches prior experience, which the solver then refines. For rotation tasks, cold starts begin with linear and suboptimal orientation trajectories, while warm starts begin smoother and closer to the optimal solution, providing a better initialization. 

Importantly, in all cases the final trajectory is the same, reaffirming that the optimization dominates the solution quality. However, the warm start initialization provides a better starting point, reducing the number of iterations required by the solver and improving computational efficiency.

\textbf{Implications for Onboard and Adaptive Planning:} 
These results have important implications for autonomy algorithms in time- and compute-constrained settings, such as space robotics. Learned warm starts can significantly reduce computation time for complex planning tasks, particularly when dynamics or obstacles introduce non-convexity. The results showed a 50-60\% reduction in number of iterations required for convergence, which can enable faster re-planning and unlock real-time operations in these resource-constrained environments. 

More broadly, these results highlight the promise of adaptive, learning-guided planners that accelerate optimization in structured environments while remaining robust to distributional shifts. By training on diverse environments and incorporating more expressive architectures~\cite{BridenGurgaEtAl2025b, GuffantiGammelliEtAl2024}, future systems could generalize more effectively to unseen scenarios. Augmenting these approaches with out-of-distribution (OOD) detection~\cite{BanerjeeSharmaEtAl2022,SinhaSchmerlingEtAl2023} would enable planners to fall back to cold starts or alternative initializations when model confidence is low, ensuring that reliance on learned priors is modulated intelligently based on context. 
This points toward a new generation of planners that dynamically balance learned priors for initialization with fallback mechanisms to maintain reliability in dynamic and uncertain environments.

\section{Conclusion} 
\label{sec:conclusion}
In this work, we presented the first flight results on the use of machine learning to warm start and significantly accelerate runtimes for onboard trajectory optimization for the Astrobee free-flying robot.
Through our experimental tests onboard the~\gls{iss} in February, 2025, we demonstrated how our trained network reduces the number of solver iterations required for convergence in complex, non-convex planning tasks, including a 60\% iteration reduction for convergence in scenarios involving rotational dynamics and a 50\% reduction in cases with obstacles drawn from the training distribution of the warm start model.
Through this effort, we aim to demonstrate how machine learning can be safely infused for onboard guidance, navigation, \& control and unlock frontiers in autonomous capabilities for the next generation of spaceflight missions.

\vspace{-0.3em}

\small
\section*{Acknowledgments}
We thank all the supporting partners from the Astrobee group at NASA Ames Research Center, the~\gls{iss} program at NASA Johnson Space Center, the NASA Space Technology Mission Directorate, and the Office of Naval Research.
This research was supported in part under the NASA Early Stage Innovation grant NNX16AD19G and NASA Space Technology Graduate Fellowship grant NNX16AM78H. We would like to thank Abla Ghaleb and Isabella Torres for their work on earlier implementations of the GuSTO trajectory optimization algorithm.
We would also like to thank the Astrobee operations team at NASA Ames, including Henry Orosco, Jose Benavides, Aric Katterhagen, Jonathan Barlow, Cristian Garcia and especially Ruben Garcia Ruiz, who supported multiple ground testing sessions, assisted with software questions and library cross-compilation (including the first-ever cross-compilation of PyTorch for Astrobee), and generally made this research possible. 

\bibliographystyle{IEEEtran}
\bibliography{bibs/ASL,bibs/main,bibs/project}

\end{document}